\documentclass{egpubl}
\usepackage{pg2024}
\makeatletter
\EGlocalpagenumber
\copyrightTextTitPag{\copyright\ 2024\ The Authors.\\
Proceedings published by Eurographics - The European Association for Computer Graphics. \\
This is an open access article under the terms of the Creative Commons Attribution License, which permits use, distribution and reproduction in any medium, provided the original work is properly cited.}
\def\@oddfoot{{\tiny\raisebox{\z@}[8pt][1pt]{\parbox[t]{20pc}{\sloppy}}}}
\def\@evenfoot{\mbox{}\hfill {\tiny\raisebox{\z@}[8pt][1pt]{\parbox[t]{20pc}{\sloppy}}}}
\def\ps@titlepage{\let\@mkboth\@gobbletwo
 \def\@oddhead{\raisebox{\z@}[8pt][1pt]{\parbox{\textwidth}{\small\parbox[t]{.7\textwidth}{\sloppy\raggedright
     Pacific Graphics 2024 /  R.\ Chen, T.\ Ritschel, and E.\ Whiting}
   \hfill\parbox[t]{.3\textwidth}{\sloppy\raggedleft
   \textit{Conference Paper}}}%
 }%
}
 \def\@oddfoot{{\tiny\raisebox{\z@}[8pt][1pt]{\parbox[t]{20pc}{\sloppy
  \p@copyrightTextTitPag}}}\hfill}
}
\makeatother

\CGFStandardLicense

\usepackage[T1]{fontenc}
\usepackage{dfadobe}  
\usepackage{amssymb} 
\usepackage{amsmath} 
\usepackage{amssymb} 
\usepackage{multirow} 
\usepackage{array} 
\usepackage{arydshln} 

\usepackage{cite}  
\BibtexOrBiblatex
\electronicVersion
\PrintedOrElectronic
\ifpdf \usepackage[pdftex]{graphicx} \pdfcompresslevel=9
\else \usepackage[dvips]{graphicx} \fi

\usepackage{egweblnk} 

\title[3DStyleGLIP: Part-Tailored Text-Guided 3D Neural Stylization]%
      {3DStyleGLIP: Part-Tailored Text-Guided 3D Neural Stylization}

\author[SeungJeh Chung \& JooHyun Park \& HyeongYeop Kang]
{\parbox{\textwidth}{\centering 
SeungJeh Chung $^{1}$\orcid{0009-0000-5306-6896},
JooHyun Park $^{1}$\orcid{0009-0009-5891-7405}, and
HyeongYeop Kang $^{2, *}$\orcid{0000-0001-5292-4342}
        }
        \\
{\parbox{\textwidth}{\centering $^1$Kyung Hee University, South Korea
         $^2$Korea University, South Korea
       }
}
}

\begin{document}

\teaser{
 \includegraphics[width=0.9\linewidth]{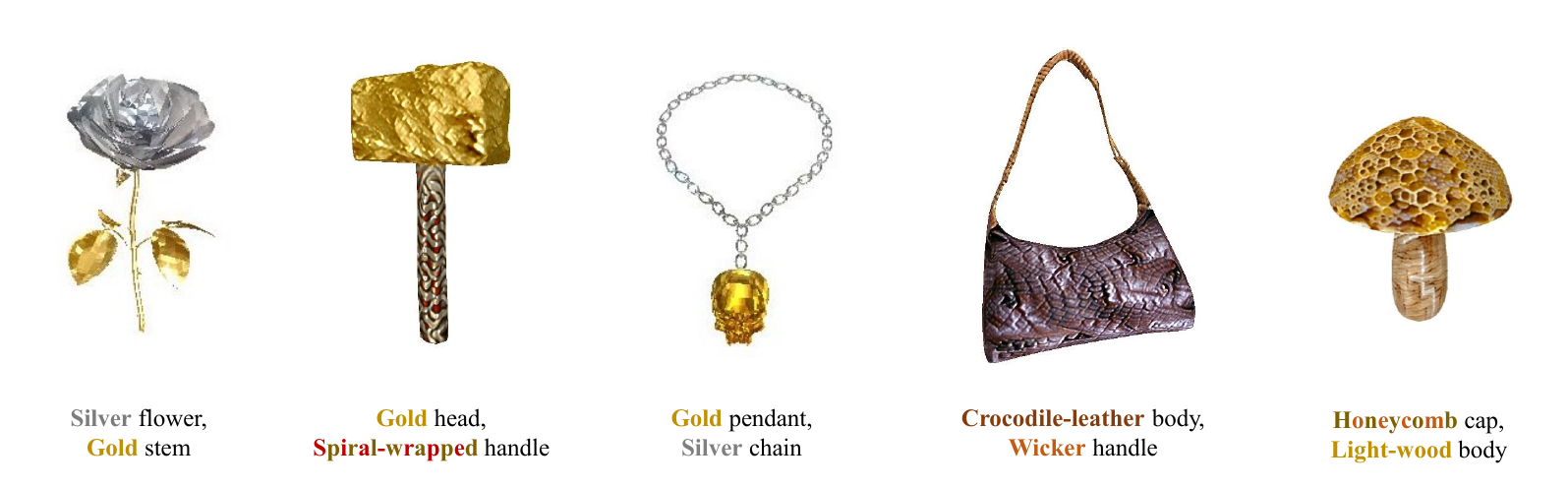}
 \centering
  \caption{Selected examples of part-tailored stylization executed by 3DStyleGLIP, demonstrating the framework's exceptional versatility in adapting diverse styles to individual components of various 3D models. This presents the framework's advanced capabilities in achieving detailed, contextually plausible stylizations.}
\label{fig:teaser}
}

\maketitle
\begin{abstract}
3D stylization, the application of specific styles to three-dimensional objects, offers substantial commercial potential by enabling the creation of uniquely styled 3D objects tailored to diverse scenes. Recent advancements in artificial intelligence and text-driven manipulation methods have made the stylization process increasingly intuitive and automated. While these methods reduce human costs by minimizing reliance on manual labor and expertise, they predominantly focus on holistic stylization, neglecting the application of desired styles to individual components of a 3D object. This limitation restricts the fine-grained controllability. To address this gap, we introduce 3DStyleGLIP, a novel framework specifically designed for text-driven, part-tailored 3D stylization. Given a 3D mesh and a text prompt, 3DStyleGLIP utilizes the vision-language embedding space of the Grounded Language-Image Pre-training (GLIP) model to localize individual parts of the 3D mesh and modify their appearance to match the styles specified in the text prompt. 3DStyleGLIP effectively integrates part localization and stylization guidance within GLIP's shared embedding space through an end-to-end process, enabled by part-level style loss and two complementary learning techniques. This neural methodology meets the user's need for fine-grained style editing and delivers high-quality part-specific stylization results, opening new possibilities for customization and flexibility in 3D content creation. Our code and results are available at https://github.com/sj978/3DStyleGLIP.


\end{abstract}  
\section{Introduction}
\label{sec:introduction}

3D stylization is a sophisticated technique that applies a specific style to three-dimensional objects by tailoring the visual attributes such as shape, texture, and color of three-dimensional objects.
It can enhance visual narratives in gaming, metaverse, and cinema by enabling the creation of diverse 3D objects with distinct moods and styles, tailored to specific thematic and aesthetic demands of different scenes. 
Such versatility in design is achieved without necessitating the creation of each object from scratch, which inherently streamlines the production workflow.

Despite its potential, 3D stylization has traditionally been underutilized, primarily due to the necessity of skilled designers for effective stylization.
However, recent advancements in machine learning have revolutionized the realm of 3D stylization, paving the way for automating stylization without the necessity of skilled professionals. 
This technique, called 3D neural stylization, employs advanced algorithms to guide the stylization process using various style cues, including reference images and textual descriptions.

A significant advancement in this domain has been the development of Vision-Language Pre-Trained Models~\cite{du2022survey}, adept at correlating images with corresponding text descriptions. These models have unlocked the potential for achieving high-quality stylization outcomes based on text prompts that describe the desired aesthetics.
For example, Text2Mesh~\cite{michel2022text2mesh} leverages CLIP's embedding space~\cite{radford2021learning} to modify 3D meshes that represent styles delineated by text inputs. TANGO~\cite{chen2022tango} later expands this work by incorporating reflectance properties, local geometric variation, and lighting conditions to generate more photorealistic stylization results. Moreover, recent methods based on diffusion models~\cite{richardson2023texture, yang20233dstyle}, known for improving fine-grained details and visual quality, also draw upon CLIP's cross-modal understanding.

However, these CLIP-based stylization methods have an inherent limitation: they allow \textit{holistic stylization}, applying a consistent style across an entire mesh, but fall short in achieving \textit{part-tailored stylization}, where distinct styles are applied to different parts of the mesh.
For instance, while CLIP-based methods can transform an entire hammer into a wooden design, they struggle to apply different styles to separate components like a striped leather handle and a cracked stone head. 
This limitation stems from the inherent design of CLIP, which learns a global-level neural representation rather than a local-level one. Such an inability becomes a more pronounced drawback in practical and industrial use, where diverse styles across different regions of a mesh are typically required.

In response to this challenge, we introduce 3DStyleGLIP, a novel framework that facilitates part-tailored 3D neural stylization, significantly advancing the controllability of fine-grained stylization.
Given text prompts comprising style and part phrases (e.g., ``pottery base, gold tube, linen shade" for lamp object), 3DStyleGLIP first identifies the corresponding parts in a given 3D mesh.
It then simultaneously modifies the mesh's local geometric details and colors to align with the specified style.

The foundation of this method lies in leveraging GLIP~\cite{li2022grounded}, a vision-language pre-trained model for object detection and phrase grounding, providing local-level neural representation. 
By inputting 2D images rendered from multiple views of the 3D scene and corresponding text prompts into GLIP, the model localizes the target parts specified by the part phrases. It estimates the similarity between these localized regions and the style phrases within GLIP’s vision-language embedding space. 
This similarity estimation guides learnable stylization modules, including the neural fields and spherical Gaussian functions as described in \cite{chen2022tango}, which are designed to parameterize the reflectance properties, local geometric variation, and lighting conditions, thereby enabling our framework to support part-tailored stylization.
Note that, we define style through physical rendering properties and guide it with GLIP to align with text prompts, achieving high-quality results comparable to diffusion-based models while maintaining style controllability.

The innovation of 3DStyleGLIP lies in its streamlined learning strategy, which unifies part localization and stylization guidance within a shared embedding space. This integrated approach fosters more stable and consistent learning outcomes. 
When part-tailored stylization relies on separate networks for localization and stylization, additional bridging is required during training. 
This bridging process typically involves data transformation and normalization steps, which frequently introduce noise and hinder stable learning. 

A summary of our contributions is as follows:
\begin{itemize}
  \item We introduce 3DStyleGLIP, a novel framework designed for text-guided, part-tailored 3D neural stylization, significantly advancing the controllability of fine-grained stylization. 
\item We introduce a joint learning of localization and stylization within GLIP's embedding space. This streamlines the learning process, facilitating stable and reliable part-tailored 3D stylization. 
  \item We devise two complementary learning techniques: a text prompt and multi-view fine-tuning technique, and an alternating training technique. These improve GLIP's ability in localization and vision-language understanding, thereby elevating the overall quality of 3D stylization.
  \item Through comprehensive experimentation, we have demonstrated that 3DStyleGLIP produces visually plausible stylization results while supporting part-tailored stylization. 
\end{itemize}

\section{Related works}
\label{sec:related works}

\subsection{3D Neural Stylization}
3D neural stylization, utilizing neural networks, manipulates the color, texture, and shape of the 3D digital representations. This facilitates the automated generation of stylized content in computer graphics. 
Recently, numerous approaches have adopted large pre-trained models such as VGG~\cite{simonyan2014very} and CLIP~\cite{radford2021learning} for efficient zero-shot feature extraction, crucial for guiding stylization via images or texts. 
The capabilities of the large pre-trained models have inspired diverse methods for stylizing 3D data in formats like volumes~\cite{kim2019transport, kim2020lagrangian, guo2021volumetric, aurand2022efficient}, implicit fields~\cite{huang2021learning, chiang2022stylizing, huang2022stylizednerf, chen2022upst, liu2023stylerf}, and point clouds~\cite{cao2020psnet}. 
The advent of a differentiable renderer~\cite{kato2018neural} has further extended neural stylization to meshes, a prevalent format used in computer graphics and 3D applications. 
Various studies~\cite{liu2018paparazzi, hollein2022stylemesh} have explored manipulating style features using images as reference, while others~\cite{chen2022tango, hwang2023text2scene, gao2023textdeformer, richardson2023texture, chen2023text2tex, zhuang2023dreameditor} have employed texts as guides for altering the mesh style. 
Moreover, methods based on Neural Radiance Fields~\cite{nguyen2022snerf, bao2023sine, fischer2024nerf, jung2024geometry} have enabled the editing of appearance using text or image references, producing high-quality 3D results.
However, these methods encounter a challenge in directly manipulating individual mesh parts due to their reliance on holistic visual representations.
To overcome this challenge, our study aims to provide more granular control over the stylization process, allowing for the targeted manipulation of individual mesh parts.

\subsection{Text-driven Manipulation}
The advent of Vision-Language Pre-Trained Models~\cite{li2019visualbert, li2020oscar, li2021align, radford2021learning, singh2022flava}, has significantly influenced text-driven manipulation in computer graphics and vision tasks.
CLIP~\cite{radford2021learning}, in particular, stands out for its extraordinary zero-shot and few-shot transferability capabilities, a result of its training on extensive image-text paired datasets.
Building on the foundation laid by CLIP, numerous studies have leveraged text-image associations to resolve various computer graphics and vision tasks. 
For instance, the integration of CLIP with styleGAN~\cite{esser2021taming, yu2021vector} has led to significant advancements in image generation control~\cite{gal2022stylegan, abdal2022clip2stylegan, pinkney2022clip2latent}.
Similarly, the integration of CLIP with VQGAN~\cite{esser2021taming, yu2021vector} offers a unified approach to semantic image generation and editing~\cite{crowson2022vqgan}.
CLIP's efficacy is also seen in guiding diffusion models toward text prompts~\cite{nichol2021glide, ramesh2022hierarchical} or achieving image style transfer~\cite{kwon2022clipstyler, xu2023stylerdalle}.

Expanding beyond 2D realms, recent developments have utilized CLIP to bridge the image-text relationship into 3D spaces.
For instance, Text2Mesh~\cite{michel2022text2mesh} and Tango~\cite{chen2022tango} have extended CLIP's capabilities to create a 3D content-text relationship through neural fields and differentiable rendering. 
CLIP's application extends to generating and manipulating 3D avatars~\cite{hong2022avatarclip, canfes2023text} and shapes~\cite{mohammad2022clip}.
The integration of the CLIP encoder with the diffusion model allows semantically rich text-based 3D generation~\cite{wang2023prolificdreamer, poole2023dreamfusion, lin2023magic3d, metzer2023latent, chen2023fantasia3d, tsalicoglou2023textmesh, haque2023instruct}. 
Despite CLIP's remarkable abilities in image-text transferability, it falls short in providing the discrete, component-specific representations necessary for part-tailored stylization. 
Aiming for text-based local-level 3D stylization, 3D Paintbrush~\cite{decatur20243d} has proposed. However, it can only stylize one semantic local region at a time and is more responsive to clothing or accessory parts rather than general parts such as the head, body, limbs, etc. Therefore, to achieve a fully stylized mesh, separate neural networks must be trained for each part, and their results must be merged. This process requires significant additional time and often results in untextured empty areas.
To address this issue, we have chosen to utilize GLIP~\cite{li2022grounded}, a vision-language model distinguished by its ability to offer local-level representations, allowing us to learn all semantic parts at once.

\subsection{Object-level Recognition}
Conventional object detection models~\cite{lin2017feature, lin2017focal, zhang2020bridging, tian2020fcos, wang2022internimage, xu2022pp, wang2022yolov7, fang2022eva} aim to identify the class and location of each object in a scene. This is achieved by learning a set of object features from large datasets such as COCO~\cite{lin2014microsoft}, PASCAL VOC~\cite{dong2021location}, Objects365~\cite{shao2019objects365} and OpenImages~\cite{kuznetsova2020open}. 
However, these models sometimes struggle to recognize objects that rarely appear or are absent in the training dataset. 
To overcome this, zero-shot detection has been introduced.
For zero-shot detection, ViLD\cite{gu2021open} distills knowledge from CLIP and ALIGN~\cite{li2021align}, both vision-language pre-trained models, into a two-stage detector.
MDETR\cite{kamath2021mdetr} and GLIP\cite{li2022grounded} train end-to-end object detection models using the multi-modal dataset (image-text pairs), achieving an explicit alignment between textual phrases and objects within images.
These models possess semantic-rich and language-aware properties, rendering them effective for tasks that require a contextual understanding of objects and comprehension of the cross-modal relationships between images and language.


\begin{figure}[t]
    \centering 
    \includegraphics[width=\columnwidth]{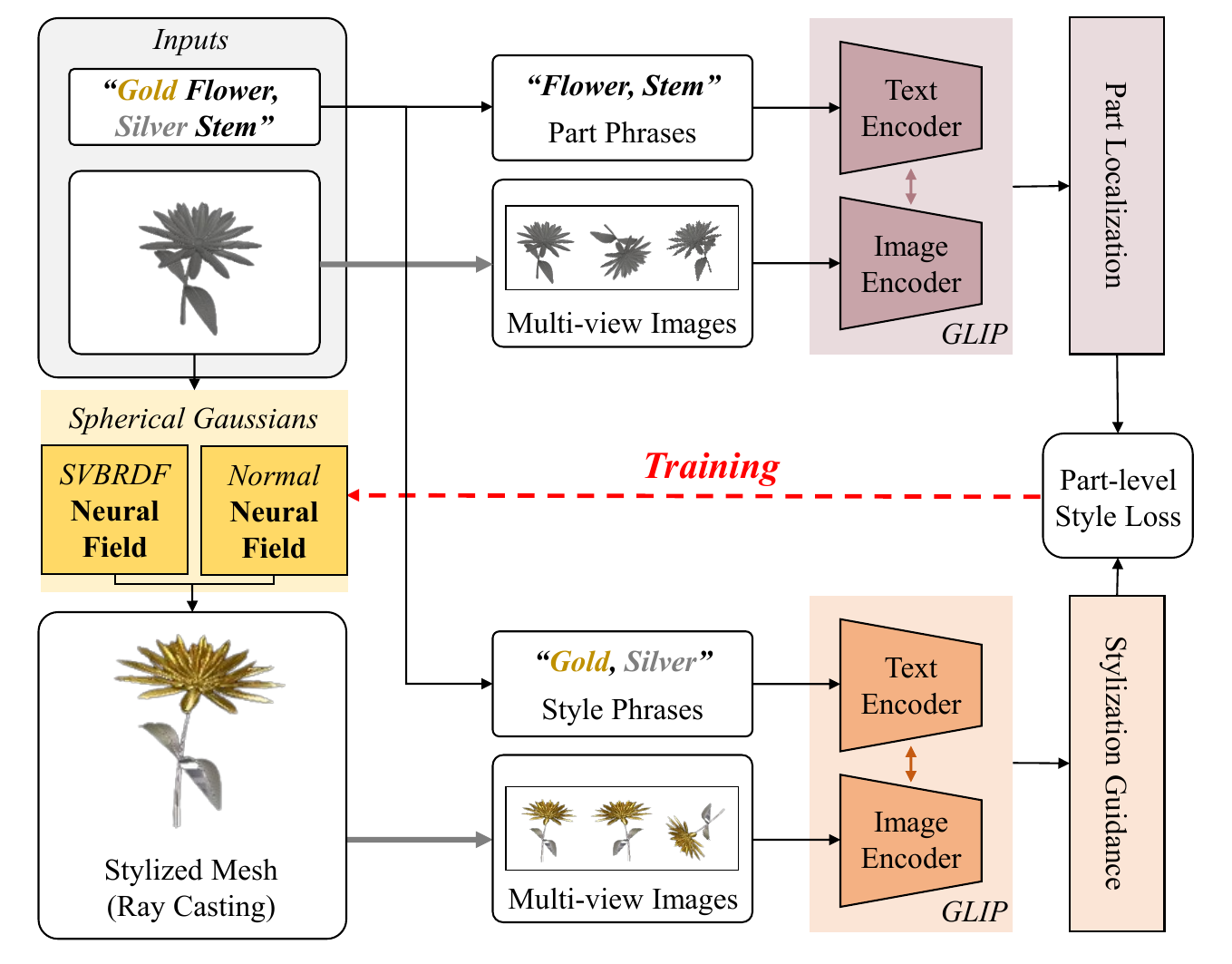} 
    \caption{The 3DStyleGLIP Pipeline. Beginning with the input of text prompts and a 3D mesh, the neural field and spherical Gaussian functions for SVBRDF, normal, and lighting conditions are trained to apply specific styles to individual parts of the 3D mesh.}
    \label{fig:pipeline}
\end{figure}

\section{Method}
\label{sec:method}
The 3DStyleGLIP framework, depicted in~\autoref{fig:pipeline}, is a text-guided, part-tailored 3D stylization system utilizing the GLIP embedding space. 
Typically, stylization tasks involve two key elements: \textit{content} and \textit{style}. 
\textit{Content} refers to a shape structure of the 3D object that is expected to be preserved after the stylization has been applied. Therefore, it is fixed during training. 
It is represented as an explicit triangle mesh $M^c$, which is composed of $N$ parts, each defined by vertices $V \in \mathbb{R}^{e \times 3}$ and faces $F \in \{1, \dots, e\}^{u \times 3}$, where $e$ and $u$ denote the number of vertices and faces, respectively.

On the other hand, \textit{style} refers to the aesthetic features such as color and local geometry that are applied to the \textit{content}. These features are modifiable based on a target text prompt $T$, which directs our training process.
It is disentangled as reflectance properties and scene lighting, which are defined as three style parameters: spatially varying BRDF (SVBRDF), normal, and lighting properties, respectively \cite{chen2022tango}.
Given a set of style phrases $Ph^{s}=\{ph^s_{1}, ph^s_{2}, \dots, ph^s_{N}\}$ and a set of part phrases $Ph^{c}=\{ph^c_{1}, ph^c_{2}, \dots, ph^c_{N}\}$, $T$ is formulated with a set of phrasal pairs.
For example, when $Ph^s$ = ``gold-chain, brown-leather" and $Ph^c$ = ``handle, body", the prompt $T$ becomes ``gold-chain handle, brown-leather body."

\begin{figure*}[t]
    \centering 
    \includegraphics[width=\textwidth]{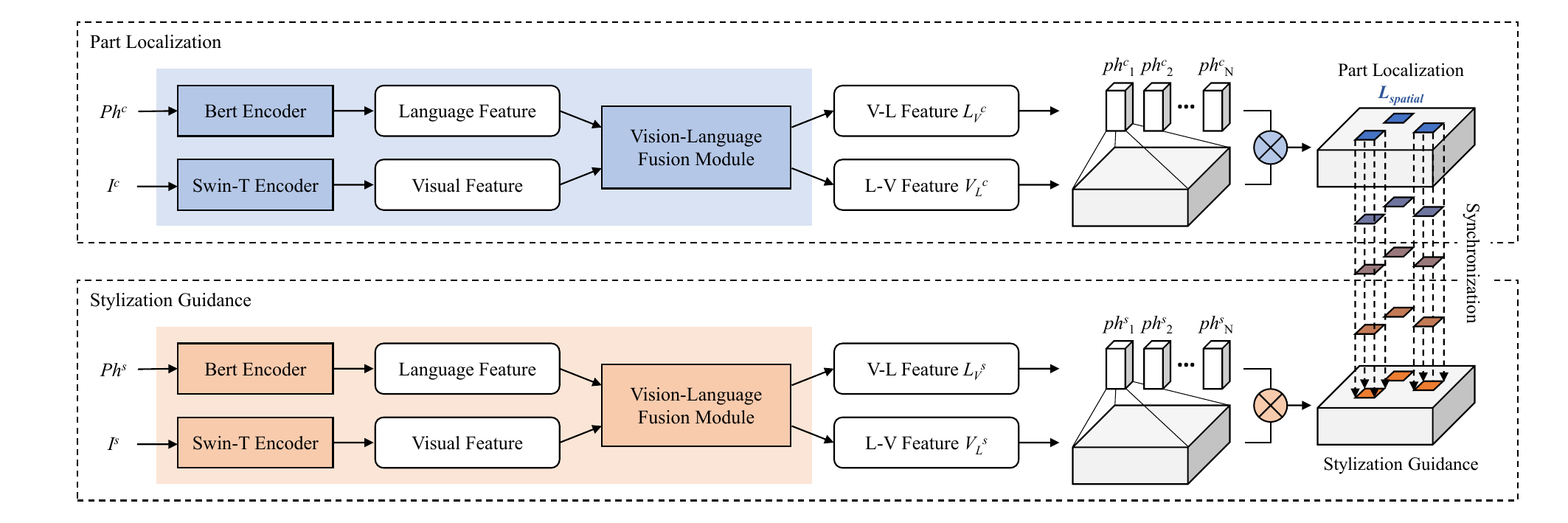}
    \caption{The joint training of part localization and stylization guidance in a shared embedding space.}
    \label{fig:GLIP process}
\end{figure*}

\subsection{Forward Process: Stylization and Rendering}
\label{subsec:forward}

Given the mesh $M^c$, it is initially scaled to fit within a unit sphere. Then, camera positions are sampled on the sphere. This begins by selecting a strategically chosen \textit{anchor viewpoint}, identified by the highest average box confidence score from GLIP. Subsequently, two additional viewpoints are selected using Gaussian random sampling centered around the \textit{anchor viewpoint}. 
For each sampled camera position $c$ and a pixel $p$ in the rendered image, we formulate a camera ray $R_p=c+tv_p$, where $v_p$ denotes the direction vector from $c$ to $p$. 
The ray casting method is then used to find the first intersection point $x_p$ and face $f_p$ on mesh $M^c$ along the ray $R_p$.

Based on the $x_p$, we employ the neural field, a coordinate-based neural network, efficiently learning high-resolution 3D objects without compromising significant computational resources~\cite{michel2022text2mesh}.
The neural field processes 3D coordinates to predict field quantities like radiance, color, and signed distance~\cite{jing2019neural, mildenhall2021nerf}.
In our configuration, the neural field encodes 3D coordinates via positional encoding and the learnable MLP networks for SVBRDF and normal.
Firstly, an SVBRDF neural field is used to estimate the surface reflectance coefficients $f_r(v_p, \omega_i, x_p)$ of the material at location $x_p$ from the viewing direction $v_p$ and the incident light direction $\omega_i$. Starting with a common 2-layer MLP, it then branches into three separate 3-layer MLPs designed to predict the specular reflectance, diffuse albedo, and roughness, respectively.
Next, a Normal neural field is utilized to predict the normal vector $\hat{n}_p$ of $x_p$ according to the normal vector $n_p$ of $f_p$. In practice, the normal at $x_p$ is calculated by predicting the offset of $n_p$, which is obtained through a 4-layer MLP.
On the other hand, light properties are designed using multiple spherical Gaussians (SG) functions to estimate incident light intensity $L_i(\omega_i)$. Each SG function models light intensity focused in a specific direction, enabling a simple and efficient representation of complex lighting distributions. By combining multiple SG functions, we can accurately approximate the overall lighting effects in a scene.
Accordingly, the final pixel color is represented as the observed light intensity $L_p(v_p, x_p, n_p)$, which is an integral over the hemisphere $\Omega = \left\{ \omega_i : \omega_i \cdot \hat{n}_p \geq 0 \right\}$:
\begin{equation}
\label{eqn:01}
L_{p}(v_{p}, x_{p}, n_{p}) = \int_{\Omega} L_{i}(\omega_{i}) f_{r}(v_{p}, \omega_{i}, x_{p})(\omega_{i} \cdot \hat{n}_{p}) d\omega_{i}.
\end{equation}
For more detailed information about the modeling of each style parameter, please refer to \cite{chen2022tango}.

To render an image, a set of rays is sampled corresponding to all the pixels in the image, and the resulting colors are assembled into the final image. We employ an SG differentiable renderer, which facilitates the backpropagation of gradients into the three learnable style parameters.
The reflectance properties and scene lighting, which are estimated by the learned neural fields and SG functions, transform colors and geometry details from the given mesh $M^c$ to the stylized mesh $M^s$, reflecting the style defined by $T$. 
To achieve this, we train the style parameters by minimizing the distance within the GLIP embedding space between the rendered images of $M^s$ and their corresponding style phrases within $T$.
Due to the prediction of style parameters for every intersection points of the camera ray, the quality of the mesh, such as triangulation and smoothness, does not significantly affect the result~\cite{chen2022tango}.

\subsection{Part-level Style Loss using GLIP}
\label{subsec:styloss}
The neural fields and SG functions training in 3DStyleGLIP involve part localization and stylization guidance, as depicted in~\autoref{fig:GLIP process}. 
During part localization, mesh $M^c$ is rendered into an image $I^c$. GLIP then processes $I^c$ and part phrases $Ph^c$, encoding them into its vision-language embedding space using separate language and image encoders.
Following this, a vision-language fusion module takes those two types of features as input, yielding language-aware visual features $V_L^c \in \mathbb{R}^{w \times h \times d}$ and visual-aware language features $L_V^c \in \mathbb{R}^{N \times d}$, where $w$ is the width, $h$ is the height, $d$ is the feature hidden dimension. 

Then, $V_L^c$ is utilized by a part detection module to predict the positions and sizes of constituent parts in $M^c$, which are represented by 2D bounding boxes. 
Furthermore, a region-word alignment score $S_{align}$ is calculated from $V_L^c$ and $L_V^c$ to estimate the correspondence between localized regions in the image and their corresponding part phrases. 
Let $E^G_{I}$ denote the GLIP's image encoder, and $E^G_{L}$ denote the GLIP's language encoder. Then, $S_{align}$ is computed as follows:
\begin{equation}
\label{eqn:01}
S_{align}=V_L^c(L_V^c)^T, V_L^c=E^G_{I}(I^c, Ph^c), L_V^c=E^G_{L}(I^c, Ph^c),
\end{equation}
When $S_{align}$ exceeds a predefined threshold (0.5 in our implementation), the region is deemed to be successfully localized, defining a spatial location set $L_{spatial}=\{(x_{1}, y_{1}), (x_{2}, y_{2}), \dots, (x_{K}, y_{K})\}$ for the part features, where $K$ is the number of positions of local region that aligned with part phrases.
Each tuple $(x_{i}, y_{i})$ represents the 2D position within the GLIP embedding space of the image, as shown in the last step in~\autoref{fig:GLIP process}.

After part localization, $M^s$ are rendered into an image $I^s$, using the same viewpoints as used in $I^c$.
Then $I^s$ and $Ph^s$ are input into GLIP to project them into the vision-language embedding space. 
To apply distinct styles, each part specified by $Ph^s$ must be identified within $I^s$.
Since $M^s$ and $M^c$ share the same spatial locations, $L_{spatial}$ can be used to locate part regions in $I^s$ and we call this synchronization.  
The region-word alignment score $S^{L}_{align}$ between GLIP's visual embedding of $I^s$ within $L_{spatial}$ and the GLIP language embedding of $Ph^s$ is maximized to train the neural fields and SG functions, enhancing the correspondence between localized regions and style phrases.
The neural fields and SG functions then act as a part-tailored stylization module, transforming $M^c$ to $M^s$. 

Therefore, the part-level style loss $\mathcal{L}_{ps}$ is defined to maximize $S^{L}_{align}$ as follows:
\begin{equation}
\label{eqn:02}
S_{align}=V_L^s(L_V^s)^T, V_L^s=E^G_{I}(I^s, Ph^s), L_V^s=E^G_{L}(I^s, Ph^s),
\end{equation}
\begin{equation}
\label{eqn:03}
S^{L}_{align}=\{S_{align}(x, y)|(x, y) \in L_{spatial}\},
\end{equation}
\begin{equation}
\label{eqn:04}
\mathcal{L}_{ps}=distance(\sigma(S^{L}_{align}), T_{L}),
\end{equation}
where $\sigma()$ denotes the sigmoid function, $distance()$ denotes the L2 norm, and $T_{L}$ denotes the learning target which is set to 1s, corresponding to the desired values that elements of $S^{L}_{align}$ should achieve.
It is worth noting that the $distance()$ function can be implemented using various error measurement methods such as mean squared error, binary cross-entropy, or negative mean methods. No significant performance differences were observed between these methods in our internal tests.  

\subsection{Text Prompt and Multi-View Fine-tuning}
\label{subsec:finetuning}

The process of part-tailored stylization in our 3DStyleGLIP framework initiates with GLIP's ability to localize the positions of parts in a 3D object using textual phrases. 
To adapt GLIP from its original 2D image domain to the 3D object domain, we conduct a fine-tuning to retrain the model and update its parameters.
Our fine-tuning, inspired by GLIP's easy prompt tuning and impressive zero-shot transfer learning capabilities, involves using multi-view rendered images and corresponding text prompts to address two critical aspects. 

The first aspect centers on the need for rendered images from various viewpoints of the 3D object to prominently feature the distinctive characteristics of each part.
GLIP's localization performance depends on the presence of these distinctive features in the rendered images. 
However, challenges arise with views randomly selected for localization, as they may not effectively capture the visual features that correspond to the texture phrases, such as in cases where objects like lamps or cups might only show a simple circular shape from the top or bottom views.

\begin{figure}[t]
    \centering 
    \includegraphics[width=\columnwidth]{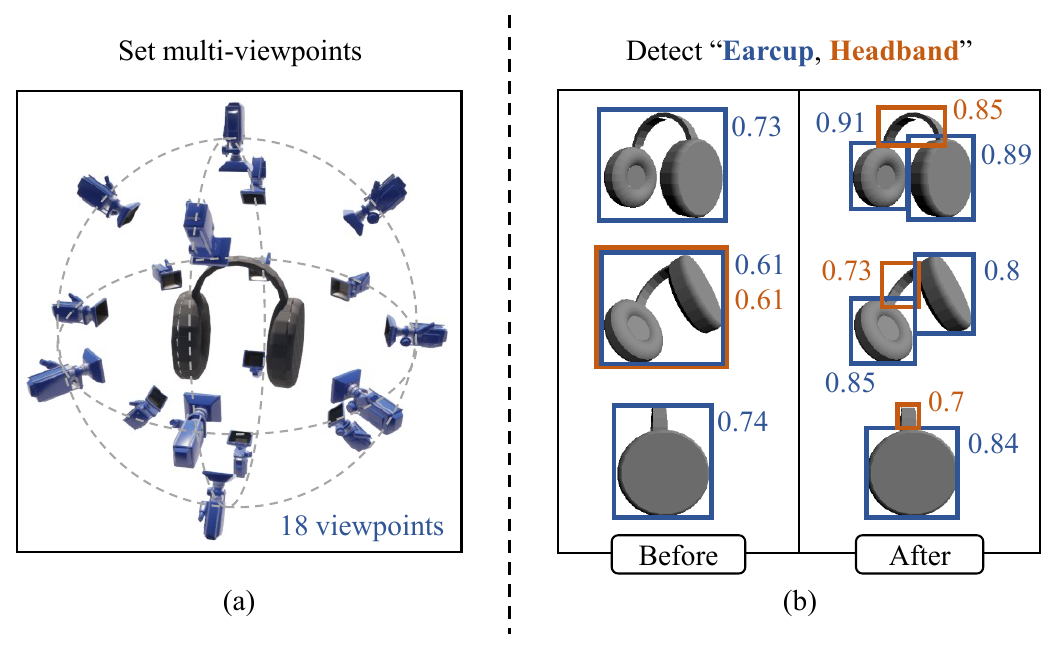}
    \caption{The visualization of the multi-view fine-tuning. (a) Uniformly distributed multiple viewpoints. (b) The comparison of GLIP Performance before and after the fine-tuning. The floating-point numbers indicate the detection accuracy of each bounding box, ranging from 0 to 1.}
    \label{fig:Fine-tune}
\end{figure}

To mitigate this, we establish uniformly distributed viewpoints to capture parts from various angles, as shown in~\autoref{fig:Fine-tune}. 
For each viewpoint, bounding boxes corresponding to the part phrases in the text prompts are generated and used as ground-truth labels.
When a target appears too small or indistinct, its bounding box is omitted to ensure the dataset's plausibility.
This fine-tuning results in GLIP achieving an improved part detection capability in the 3D mesh domain, with an Average Precision (AP) score between 0.8 and 0.9. 

The second aspect deals with the variability in textual expressions referring to the same object parts.  
For example, various terms like “spout,” “nozzle,” and “mouth” might refer to the same part of a kettle mesh. 
For practical use, GLIP should produce consistent localization results for these varied expressions. 
To address this, we incorporate various expressions for object parts, such as ``handle", ``grip" and ``holder" for a cup or ``torso" and ``body" for a horse, during the viewpoint fine-tuning.

Retraining and updating the model are achieved by concatenating a learnable offset vector to the language embeddings extracted from the language encoder.
This approach significantly saves computational resources required for training, focusing updates on language embeddings rather than the entire parameter set of GLIP. 
Consequently, fine-tuning is typically completed in under five minutes on a single Nvidia RTX 3070 GPU.

\begin{figure*}[t]
    \centering 
    \includegraphics[width=\textwidth]{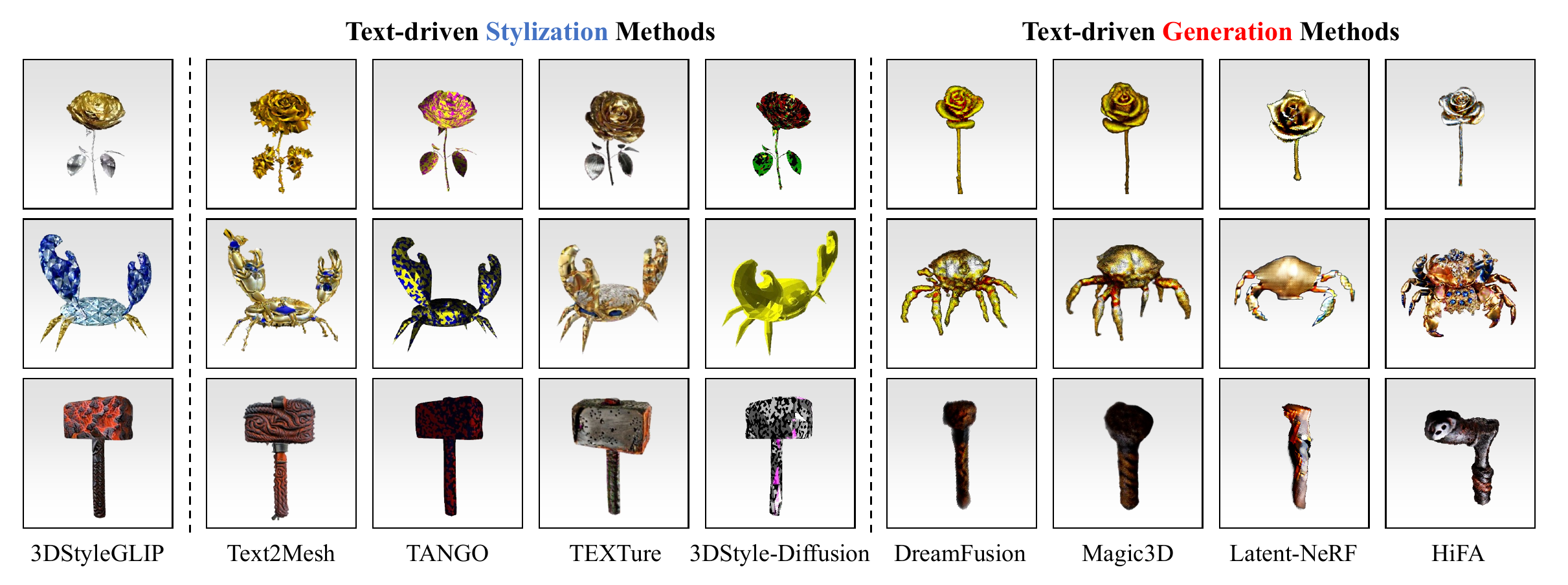}
    \caption{Comparative visualization of stylization results from eight different methodologies. This demonstrates each method's capability to recognize and apply distinct styles to individual parts of 3D meshes. The applied text prompts T are ``gold flower and silver stem" for the rose, ``diamond shell, gold legs, and sapphire claws" for the crab, and ``lava head and twisted-leather handle" for the hammer. To account for varying responsiveness to text prompts across methods, we tested multiple formulations and selected the optimal prompt for each method. Examples include: "a DSLR photo of a {object type} made of {T}", "an image of a {object type} made of {T}", and "{object type}, {T}".}
    \label{fig:5_comparison_holistic}
\end{figure*}

\subsection{Alternating Stylization with CLIP}
\label{subsec:alternating}

Text2Mesh is a pioneering approach that exploits CLIP's image-level neural representation for holistic 3D stylization. 
It trains the neural fields and SG functions to maximize the cosine similarity between the rendered image of a 3D object and a style prompt within the CLIP's embedding space. 
Drawing inspiration from Text2Mesh, 3DStyleGLIP focuses on aligning visual regions in the rendered image of a 3D object with corresponding style phrases, thus facilitating part-tailored stylization.

However, GLIP, despite its semantic richness learned from an extensive dataset of 3 million human-annotated and 24 million web-crawled image-text pairs, occasionally falls short in delivering satisfactory stylization results.
We speculate that this limitation arises because GLIP's formulation prioritizes vision-language localization over comprehensive vision-language understanding. 
As the updated GLIPv2 model, which is expected to augment GLIP’s vision-language understanding capabilities, remains unreleased, we integrate a supplementary technique using CLIP's robust embedding features for stylization.

To augment GLIP's stylization performance, we introduce a CLIP-based loss $\mathcal{L}^\ast$.
This process involves cropping regions from the rendered image $I^s$, which correspond to $L_{spatial}$, and generating a set of cropped images $I^s_{crop}$. These images are then projected into CLIP's embedding space using the CLIP image encoder $E^C_{I}$, while the corresponding style phrases $Ph^s$ are mapped using the CLIP language encoder $E^C_{L}$. 
The training is then geared towards maximizing $\mathcal{L}^\ast$, defined as the cosine similarity between CLIP embeddings of augmented $I^s_{crop}$ and $Ph^s$:
\begin{equation}
\label{eqn:05}
\mathcal{L}^\ast=sim(E^C_{I}(\psi(I^s_{crop})), E^C_{L}(Ph^s))
\end{equation}
where $\psi$() denotes the global and local augmentations introduced in \cite{michel2022text2mesh}, and $sim$ denotes the cosine similarity function. 

Our optimization strategy involves alternating between part-level style loss $\mathcal{L}_{ps}$ and CLIP-based style loss $\mathcal{L}^\ast$, thereby focusing on the unique contributions of each loss and facilitating the learning of their implicit interaction. 
It is noteworthy that our internal test suggests that this alternating training typically produces superior outcomes compared to the amalgamation of these two losses into a singular loss through weighted summation, with no significant performance differences. 
Furthermore, integrating alternating learning by incorporating the CLIP loss into each optimization iteration does not significantly compromise stability. 


\section{Experiments}
\label{sec:experiments}

We conduct an extensive evaluation of 3DStyleGLIP across a diverse array of object meshes and target styles. 
The object meshes are sourced from: PSB~\cite{chen2009benchmark}, COSEG~\cite{wang2012active}, modelnet40~\cite{wu20153d}, HumanSeg~\cite{maron2017convolutional}, toys4K~\cite{Toys4K}, manifold40~\cite{hu2022subdivision}, and Turbo Squid~\cite{turbosquid2022}. 

The average mesh in our experiments comprises approximately 34,551 vertices, with each mesh containing between 2 to 5 semantic parts.
These semantic parts are balanced in terms of size distribution, ranging from smaller components to larger ones, and encompass a wide spectrum of complexities, from simplistic structures like a cup's body to more intricate forms such as an animal's head. 
The training of 3DStyleGLIP are executed over 1500 iterations, taking roughly 30 minutes on a single Nvidia RTX 3090 GPU. 
High-quality stylization results are typically observable after performing around 600 iterations.

\subsection{Part-tailored Stylization Results}
\label{subsec:part-tailored stylization}
\subsubsection{Qualitative Study}
\label{subsec:qualitative study}

To demonstrate the superiority of our 3DStyleGLIP in executing part-tailored stylization, we conducted two comparative analyses.
First, we assessed the quality of 3D objects generated by our framework against leading text-driven 3D generation models. Considering the advancements in the latent diffusion model~\cite{rombach2022high}, we used four state-of-the-art methods as baselines: DreamFusion~\cite{poole2023dreamfusion}, Magic3D~\cite{lin2023magic3d}, Latent-NeRF~\cite{metzer2023latent}, and HiFA~\cite{zhu2023hifa}.
Unlike 3DStyleGLIP, these models generate entire geometries, leading to differences in object shapes.

Our second analysis focused on the part-wise texture quality, comparing 3DStyleGLIP with text-driven 3D stylization models that adjust a given mesh. 
We benchmarked against Text2Mesh~\cite{michel2022text2mesh} and TANGO~\cite{chen2022tango}, which leverage the CLIP embedding space, as well as TEXTure~\cite{richardson2023texture} and 3DStyle-Diffusion~\cite{yang20233dstyle}, which utilize latent diffusion models.

We utilized the official implementations of each method. Where official versions were unavailable, we utilized highly credible re-implementations~\cite{threestudio2023}.
In each model, the text prompts utilized are a combination of basic prompts provided in the paper, along with prompts specifying styles for individual parts.

As shown in~\autoref{fig:5_comparison_holistic}, 3D generation methods often produce noisy geometries, resulting in lower quality compared to human-made 3D objects.
These methods also struggle to handle detailed text prompts that specify distinct styles for different parts, leading to stylized meshes that fail to match the text input and exhibit unrealistic deformations.
Furthermore, both CLIP-based and diffusion-based 3D stylization approaches are unable to recognize and preserve the individuality of each part within a mesh, often causing unintended style blending across various parts.
 
\begin{figure*}[htp]
    \centering
    \includegraphics[width=\textwidth]{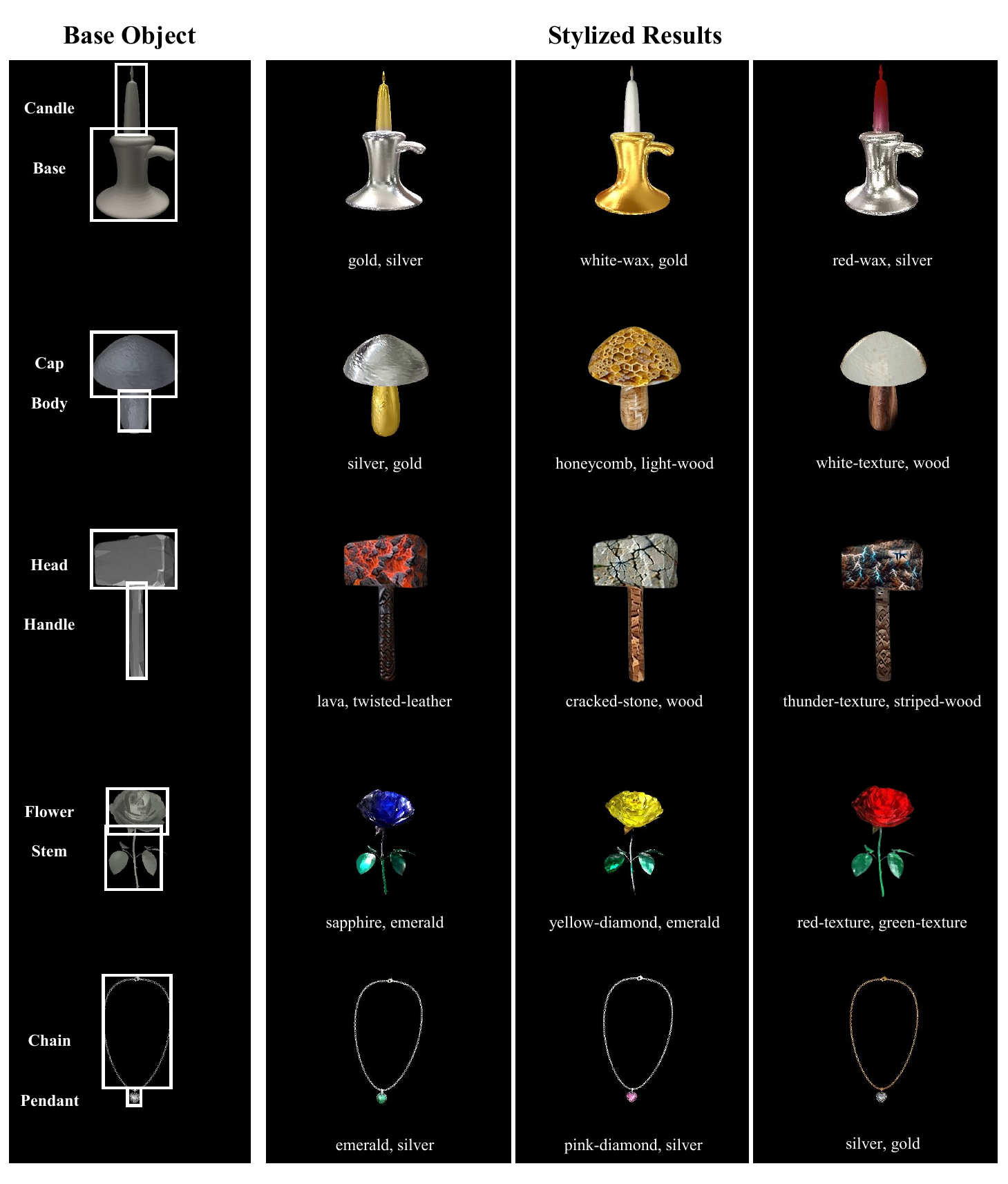}
    \caption{A variety of examples of part-tailored stylization outcomes generated by 3DStyleGLIP, illustrating the framework's capability to produce high-quality results across a range of scenarios.}
    \label{fig:all_samples}
\end{figure*}

In contrast, 3DStyleGLIP adeptly manages part-tailored stylization, consistently yielding high-quality, stylized outcomes that preserve the distinct individuality of each part. More stylization examples can be found in~\autoref{fig:all_samples}.

\subsubsection{User Study}
To demonstrate the perceptual and subjective superiority of 3DStyleGLIP compared to traditional baseline models, we conducted an extensive online study with thirty participants. This evaluation focused on the general capabilities of 3D generation and editing by selecting ten diverse 3D meshes. These meshes ranged from simple geometries, like a hammer, to complex forms, such as a crab, as shown in~\autoref{fig:5_comparison_holistic}. 
We generated ten test sets, each comprising nine test pairs, where each pair included a 3D mesh stylized by either one of eight baseline models or 3DStyleGLIP, along with corresponding text prompts.
Note that the type of stylized mesh was consistent within a test set and was different across test sets. 

Participants were randomly assigned two of these test sets. They evaluated the alignment between the stylized outputs and their respective text prompts using a 5-point Likert scale ranging from 1 (very poor) to 5 (very good).
They answered three research questions: (Q1) ``How natural does the stylized 3D object appear?" (Q2) ``How well does the appearance of the stylized 3D object align with the provided text description?" (Q3) ``How well does each part of the stylized 3D object align with the provided text description?"

\begin{table}[h]
\centering
\begin{tabular}{p{1.6cm} p{1.7cm} p{1.7cm} p{1.7cm}}
\hline
 & \centering Q1($\uparrow$) & \centering Q2($\uparrow$) & \centering Q3($\uparrow$) \tabularnewline
\hline
\centering Text2Mesh \\ ~\cite{michel2022text2mesh} & \centering\multirow{2}{*}{2.84 ($\pm$1.18)} & \centering\multirow{2}{*}{2.36 ($\pm$0.90)} & \centering\multirow{2}{*}{2.16 ($\pm$0.77)} \tabularnewline
\centering TANGO \\ ~\cite{chen2022tango} & \centering\multirow{2}{*}{1.57 ($\pm$0.73)} & \centering\multirow{2}{*}{1.57 ($\pm$0.70)} & \centering\multirow{2}{*}{1.46 ($\pm$0.84)} \tabularnewline
\hdashline
\centering TEXTure \\ ~\cite{richardson2023texture} & \centering\multirow{2}{*}{2.50 ($\pm$0.81)} & \centering\multirow{2}{*}{2.11 ($\pm$0.86)} & \centering\multirow{2}{*}{2.03 ($\pm$1.01)} \tabularnewline
\centering 3DStyle-Diffusion \\ ~\cite{yang20233dstyle} & \centering\multirow{3}{*}{1.32 ($\pm$0.52)} & \centering\multirow{3}{*}{1.21 ($\pm$0.46)} & \centering\multirow{3}{*}{1.21 ($\pm$0.46)} \tabularnewline
\hdashline
\centering DreamFusion \\ ~\cite{poole2023dreamfusion} & \centering\multirow{2}{*}{1.26 ($\pm$0.68)} & \centering\multirow{2}{*}{1.23 ($\pm$0.70)} & \centering\multirow{2}{*}{1.21 ($\pm$0.68)} \tabularnewline
\centering Magic3D \\ ~\cite{lin2023magic3d} & \centering\multirow{2}{*}{1.41 ($\pm$0.85)} & \centering\multirow{2}{*}{1.29 ($\pm$0.78)} & \centering\multirow{2}{*}{1.21 ($\pm$0.52)} \tabularnewline
\centering Latent-NeRF \\ ~\cite{metzer2023latent} & \centering\multirow{3}{*}{1.29 ($\pm$0.52)} & \centering\multirow{3}{*}{1.23 ($\pm$0.50)} & \centering\multirow{3}{*}{1.26 ($\pm$0.50)} \tabularnewline
\centering HiFA \\ ~\cite{zhu2023hifa} & \centering\multirow{2}{*}{1.95 ($\pm$0.64)} & \centering\multirow{2}{*}{1.66 ($\pm$0.70)} & \centering\multirow{2}{*}{1.52 ($\pm$0.35)} \tabularnewline
\hdashline
\centering 3DStyleGLIP \\ \textbf{(ours)} & \centering\multirow{2}{*}{\textbf{\underline{3.46 ($\pm$1.28)}}} & \centering\multirow{2}{*}{\textbf{\underline{4.25 ($\pm$1.07)}}} & \centering\multirow{2}{*}{\textbf{\underline{4.25 ($\pm$1.18)}}}
\tabularnewline
\hline
\end{tabular}
\caption{User study results conducted with thirty online participants indicate that 3DStyleGLIP surpasses the baseline methods in terms of natural appearance, alignment with text descriptions, and part-tailored capabilities in styling. The highest performance score in each research question is highlighted in underlined bold.}
\label{tab:userstudy}
\end{table}

Our findings, summarized in~\autoref{tab:userstudy}, reveal that 3DStyleGLIP generally achieves higher average ratings compared to baseline methods, demonstrating its superior performance in natural appearance, alignment with text descriptions, and part-tailored capability in styling.
Regarding the 3D generation methods, DreamFusion, Magic3D, Latent-NeRF, and HiFA, scored in the 1-point range across all questions, with HiFA receiving relatively higher scores for visual naturalness. However, HiFA's low scores in alignment with text prompts in stylization indicate its limitations in part-specific stylization.

Regarding the 3D stylization methods, Text2Mesh and TEXTure scored in the 2-point range, whereas TANGO and 3DStyle-Diffusion scored in the 1-point range. These results deviated from the initial expectations, indicating that the quality of results is influenced by more than just the differences between CLIP and latent diffusion models. 
Both TANGO and 3DStyle-Diffusion utilize reflectance properties and lighting conditions as style parameters, suggesting that improper handling of these physical rendering properties with complex text prompts can result in significant texture distortions. 

Remarkably, 3DStyleGLIP, despite not employing a diffusion-based approach, excels in delivering high-quality part-tailored stylization. This success is attributed to GLIP's localization and visual understanding capabilities. This underscores the practical advantages of our framework in achieving detailed and plausible 3D stylizations.

\subsection{Robustness and Flexibility of 3DStyleGLIP}
\label{subsec:robustness}
\subsubsection{Result Consistency and Stability}
Our part-level style loss performs part localization and stylization guidance in a shared embedding space, thereby ensuring consistency and stability in the optimization process across random seed values. 
To evaluate this aspect, we compare 3DStyleGLIP with a baseline model comprising two distinct neural networks: one for GLIP-based part segmentation and another for CLIP-based style guidance. 
3DStyleGLIP and the baseline model are respectively tasked with part-tailored stylization to produce a set of 100 stylized meshes using random seeds for the same object.
Each stylized mesh is rendered into an image from the same anchor view. 
We then compare all possible image pairs within the set, using four distinct image quality evaluation metrics: Structural Similarity Index (SSIM), Peak Signal-to-Noise Ratio (PSNR), Mean Squared Error (MSE), and Learned Perceptual Image Patch Similarity (LPIPS). 

For example, for a candle stylized with the text prompt ``gold base, silver candle," we generate 100 distinct stylized candle images using random seeds. We then create all possible pairs of these images and calculate image quality evaluation metrics for each pair, treating each image as the ground truth for the other.
These metrics allow us to assess the similarity of the stylized results, considering structural similarity, error in pixel values, and perceptual differences based on the human visual system.

\begin{table}[h]
\centering
\begin{tabular}{p{1.5cm} p{1.2cm} p{1.2cm} p{1.2cm} p{1.2cm}}
\hline
 & \centering MSE($\downarrow$) & \centering LPIPS($\downarrow$) & \centering SSIM($\uparrow$) & \centering PSNR($\uparrow$) \tabularnewline
\hline
\centering\multirow{2}{*}{baseline} & \centering 0.012 & \centering 0.082 & \centering 0.897 & \centering 19.179 \tabularnewline
                           & \centering ($\pm$3.967) & \centering ($\pm$3.686) & \centering ($\pm$7.339) & \centering ($\pm$0.442) \tabularnewline
\hline
\centering\multirow{2}{*}{3DStyleGLIP} & \centering \textbf{\underline{0.004}} & \centering \textbf{\underline{0.041}} & \centering \textbf{\underline{0.936}} & \centering \textbf{\underline{23.155}} \tabularnewline
                             & \centering ($\pm$1.297) & \centering ($\pm$1.788) & \centering ($\pm$4.888) & \centering ($\pm$1.008) \tabularnewline
\hline
\end{tabular}
\caption{3DStyleGLIP presents more consistent and stable stylization outcomes across the 100 trials compared to the baseline model, as indicated by lower MSE($\downarrow$), lower LPIPS($\downarrow$), higher SSIM($\uparrow$), and higher PSNR($\uparrow$). The highest performance score in each metric is highlighted in underlined bold.}
\label{tab:stability}
\end{table}

As shown in~\autoref{tab:stability}, 3DStyleGLIP achieves more consistent and stable stylization outcomes across the 100 trials compared to the baseline model.
Specifically, 3DStyleGLIP demonstrates lower variability in MSE and LPIPS scores, suggesting that our method is more reliable in producing visually similar stylization results across multiple trials, with less deviation from the mean.
Additionally, the SSIM and PSNR scores for 3DStyleGLIP are higher than those for baseline, indicating that the stylized outcomes produced by our method show more similar perceptual quality between trials. 

\begin{figure}[tp]
    \centering 
    \includegraphics[width=\columnwidth]{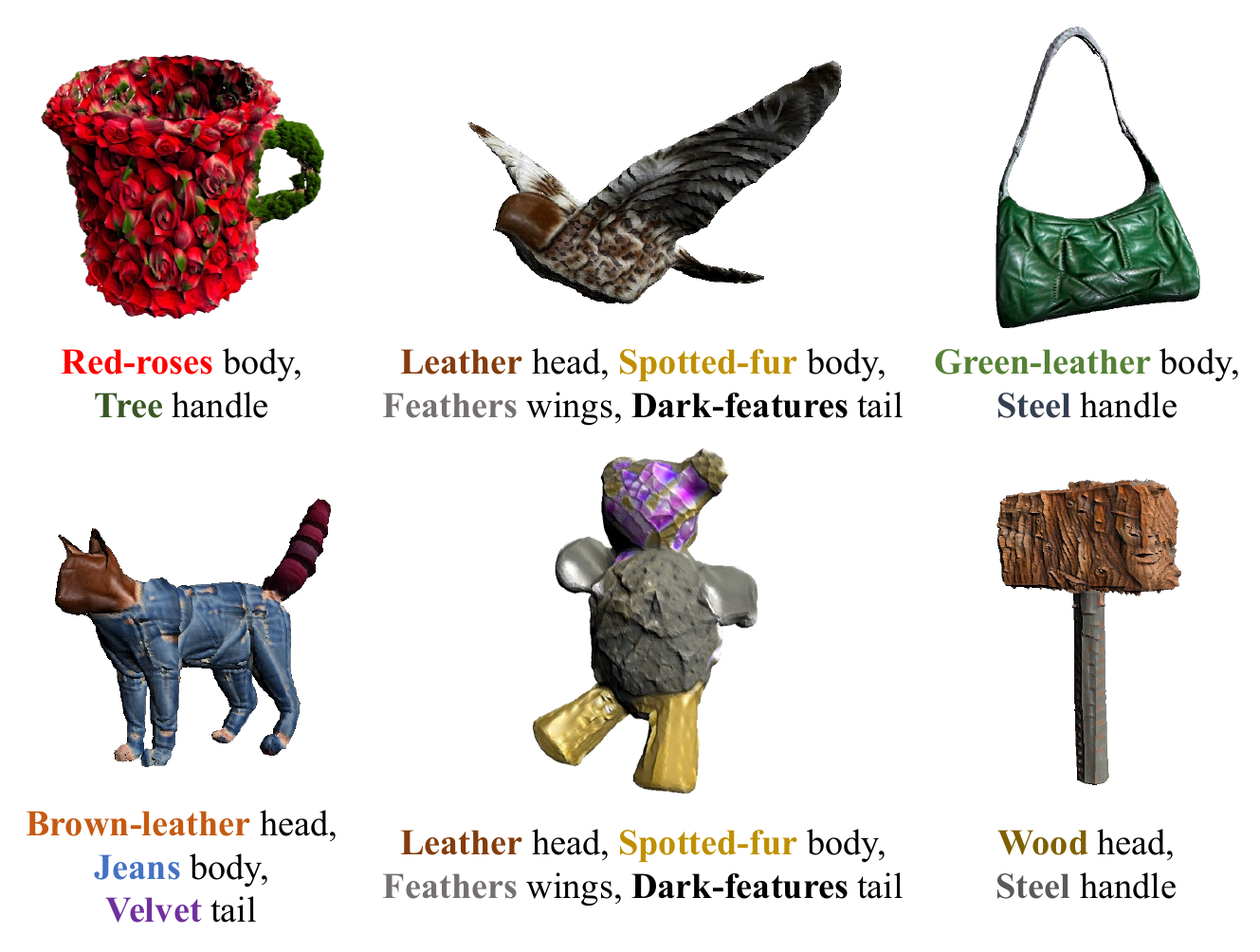}
    \caption{Neural style field is trained to predict vertex color and displacement values instead of SVBRDF, normal maps, and lighting conditions.}
    \label{fig:our_text2mesh}
\end{figure}

\subsubsection{Style Field Variant and Robustness}
In our main approach, neural fields and SG functions are trained to produce SVBRDF, normal, and lighting conditions. This process achieves photorealistic results through detailed control over reflectance and illumination. 
Thanks to the flexibility of neural fields in adapting to various field configurations~\cite{michel2022text2mesh}, 3DStyleGLIP can produce stylized textures with intricate and detailed surfaces.
In this alternative approach, we configure style field quantities to predict color and displacement values for each vertex. 
This shift necessitates a corresponding transition in the forward process from ray-casting to a differentiable rasterization method. 
In the pipeline structure of 3DStyleGLIP shown in~\autoref{fig:pipeline}, a neural style field, which is a coordinate-based MLP, is employed in place of the SG function and the two neural fields. 
The neural style field encodes vertex coordinates using positional encoding and a 4-layer MLP, which then splits into two branches that determine the color of the $i$-th vertex $c_{i} \in \mathbb{R}^3$ and displacement value of the $i$-th vertex $d_{i} \in \mathbb{R}$.

To demonstrate whether 3DStyleGLIP maintains its effectiveness in part-tailored stylization under this altered setting, we provide stylization results, as shown in~\autoref{fig:our_text2mesh}. 
The stylization outcomes are of high quality, affirming that the part-level style loss effectively ensures robust performance across a diverse range of style field configurations and objectives.

\subsection{Modules of 3DStyleGLIP}
\label{subsec:modules}
To evaluate the impact of 3DStyleGLIP's key modules on stylization tasks, we conduct an ablation study, systematically deactivating individual modules from the framework during the stylization process. 
We evaluate the performance of 3DStyleGLIP under four conditions: \textit{full}, \textit{no-GLIP}, \textit{no-finetuning}, and \textit{no-CLIP}.
In the \textit{Full} condition, 3DStyleGLIP with the full module are tested. 
In the \textit{no-GLIP}, the stylization are carried out using only the CLIP model, which is introduced for complementary stylization, without GLIP model.  
In the \textit{no-finetuning}, we exclude the fine-tuning process which are conducted for enhancing localization performance.
In the \textit{no-CLIP}, the alternating stylization, designed to enhance fine-grained style detail, are omitted.  

\begin{figure}[tp]
    \centering 
    \includegraphics[width=\columnwidth]{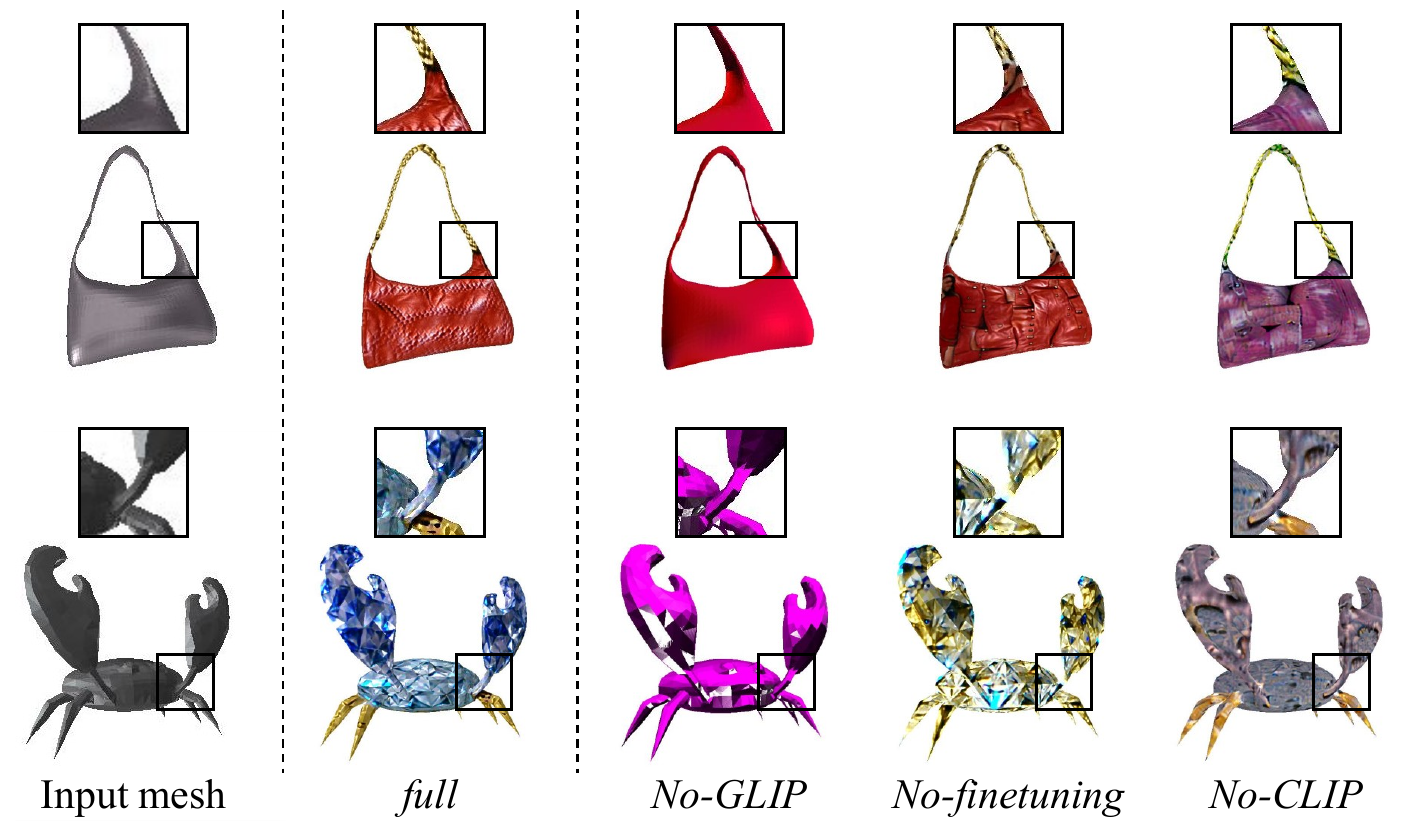}
    \caption{Ablation study on the key modules used in 3DStyleGLIP for a lamp mesh and a headset mesh. The lamp mesh is stylized with a ``wood base, gold tube, fur shade" while the headset mesh is stylized with a ``leather earcup, colorful-crochet headband". Please see \autoref{subsec:modules} for details on each condition.}
    \label{fig:Ablation}
\end{figure}

The visualization results for these test conditions are illustrated in~\autoref{fig:Ablation}.
Here, the \textit{full} serves as a baseline condition, demonstrating stylization results achievable with 3DStyleGLIP.
In terms of \textit{no-GLIP}, the result reveals a failure in object-level stylization. 
The result exhibits a single visual representation even though the text prompt provided two distinct visual representations.
This highlights the crucial role of GLIP in facilitating object-level stylization.  
In terms of \textit{no-finetuning}, the generated mesh shows a slightly mixed style. This occurred because the absence of fine-tuning caused occasional failures in recognizing mesh images from multiple viewpoints during the GLIP localization. 
In terms of \textit{no-CLIP}, while the generated mesh exhibits a distinct visual representation for each part, the applied styles seem to diverge from the intended styles. 
This occurs due to the limited VL understanding of GLIP, which poses challenges to achieving high-fidelity stylization. 
It is noteworthy that CLIP-based alternating stylization might be skipped with the adoption of the next-generation GLIP, named GLIPv2~\cite{zhang2022glipv2}. 
GLIPv2 is known for its enhanced VL understanding capabilities, as elaborated upon in GLIGEN~\cite{li2023gligen}. However, since GLIPv2 has yet to be released to the public, we will postpone further exploration in this avenue until its availability.

\section{Conclusions}
\label{sec:conclusions}
In this paper, we introduce 3DStyleGLIP, a novel framework designed for text-driven, part-tailored 3D stylization. 
As the learning of part localization and its stylization has been conducted in the GLIP's part-level vision-language embedding space, 3DStyleGLIP produces consistent and desired stylization outcomes, closely aligning with the specified text prompts. 

However, despite its demonstrated effectiveness in part-tailored 3D stylization, 3DStyleGLIP encounters certain limitations due to the intrinsic constraints of GLIP. 
Notably, while the framework proficiently manipulates specific parts of 3D objects, it currently does not support the synthesis of parts defined by texts that convey feelings or abstract concepts, such as ``delicious hamburger" or ``sad rabbit."
Moreover, the framework faces challenges in managing scenarios involving more than five object parts or those requiring the handling of highly detailed semantic parts, thereby limiting its application in complex or intricately detailed stylization tasks.

In the future, we aim to address these limitations. 
To enhance the framework's proficiency in interpreting and rendering abstract concepts and emotional contexts, we plan to incorporate advanced natural language processing techniques. 
Furthermore, to facilitate the handling of a larger number of parts or more detailed semantic parts, we intend to refine GLIP's localization performance. This will involve exploring more sophisticated segmentation algorithms and hierarchical representations, enabling finer granularity in part segmentation and stylization.
The integration of GLIPv2, the next generation of the GLIP model that will be released near future, may also be a promising solution to these problems. 

\section*{Acknowledgments}
This work was supported by ICT Creative Consilience Program through the Institute of Information \& Communications Technology Planning \& Evaluation(IITP) grant funded by the Korea government(MSIT)(RS-2020-II201819)

\bibliographystyle{eg-alpha-doi}  
\bibliography{egbibsample}        

\newcommand{\etalchar}[1]{$^{#1}$}
\begin{thebibliography}{\uppercase{MKXBP22}}

\bibitem[AONA22]{aurand2022efficient}
\textsc{Aurand J., Ortiz R., Nauer S., Azevedo V.~C.}:
\newblock Efficient neural style transfer for volumetric simulations.
\newblock \emph{ACM Transactions on Graphics (TOG) 41}, 6 (2022), 1--10.

\bibitem[AZF{\etalchar{*}}22]{abdal2022clip2stylegan}
\textsc{Abdal R., Zhu P., Femiani J., Mitra N., Wonka P.}:
\newblock Clip2stylegan: Unsupervised extraction of stylegan edit directions.
\newblock In \emph{ACM SIGGRAPH 2022 conference proceedings} (2022), pp.~1--9.

\bibitem[BZY{\etalchar{*}}23]{bao2023sine}
\textsc{Bao C., Zhang Y., Yang B., Fan T., Yang Z., Bao H., Zhang G., Cui Z.}:
\newblock Sine: Semantic-driven image-based nerf editing with prior-guided editing field.
\newblock In \emph{Proceedings of the IEEE/CVF Conference on Computer Vision and Pattern Recognition} (2023), pp.~20919--20929.

\bibitem[CADY23]{canfes2023text}
\textsc{Canfes Z., Atasoy M.~F., Dirik A., Yanardag P.}:
\newblock Text and image guided 3d avatar generation and manipulation.
\newblock In \emph{Proceedings of the IEEE/CVF Winter Conference on Applications of Computer Vision} (2023), pp.~4421--4431.

\bibitem[CBK{\etalchar{*}}22]{crowson2022vqgan}
\textsc{Crowson K., Biderman S., Kornis D., Stander D., Hallahan E., Castricato L., Raff E.}:
\newblock Vqgan-clip: Open domain image generation and editing with natural language guidance.
\newblock In \emph{Computer Vision--ECCV 2022: 17th European Conference, Tel Aviv, Israel, October 23--27, 2022, Proceedings, Part XXXVII} (2022), Springer, pp.~88--105.

\bibitem[CCJJ23]{chen2023fantasia3d}
\textsc{Chen R., Chen Y., Jiao N., Jia K.}:
\newblock Fantasia3d: Disentangling geometry and appearance for high-quality text-to-3d content creation.
\newblock \emph{arXiv preprint arXiv:2303.13873} (2023).

\bibitem[CCL{\etalchar{*}}22]{chen2022tango}
\textsc{Chen Y., Chen R., Lei J., Zhang Y., Jia K.}:
\newblock Tango: Text-driven photorealistic and robust 3d stylization via lighting decomposition.
\newblock \emph{arXiv preprint arXiv:2210.11277} (2022).

\bibitem[CGF09]{chen2009benchmark}
\textsc{Chen X., Golovinskiy A., Funkhouser T.}:
\newblock A benchmark for 3d mesh segmentation.
\newblock \emph{Acm transactions on graphics (tog) 28}, 3 (2009), 1--12.

\bibitem[CSL{\etalchar{*}}23]{chen2023text2tex}
\textsc{Chen D.~Z., Siddiqui Y., Lee H.-Y., Tulyakov S., Nie{\ss}ner M.}:
\newblock Text2tex: Text-driven texture synthesis via diffusion models.
\newblock \emph{arXiv preprint arXiv:2303.11396} (2023).

\bibitem[CTT{\etalchar{*}}22]{chiang2022stylizing}
\textsc{Chiang P.-Z., Tsai M.-S., Tseng H.-Y., Lai W.-S., Chiu W.-C.}:
\newblock Stylizing 3d scene via implicit representation and hypernetwork.
\newblock In \emph{Proceedings of the IEEE/CVF Winter Conference on Applications of Computer Vision} (2022), pp.~1475--1484.

\bibitem[CWNN20]{cao2020psnet}
\textsc{Cao X., Wang W., Nagao K., Nakamura R.}:
\newblock Psnet: A style transfer network for point cloud stylization on geometry and color.
\newblock In \emph{Proceedings of the IEEE/CVF Winter Conference on Applications of Computer vision} (2020), pp.~3337--3345.

\bibitem[CYL{\etalchar{*}}22]{chen2022upst}
\textsc{Chen Y., Yuan Q., Li Z., Xie Y. L. W. W.~C., Wen X., Yu Q.}:
\newblock Upst-nerf: Universal photorealistic style transfer of neural radiance fields for 3d scene.
\newblock \emph{arXiv preprint arXiv:2208.07059} (2022).

\bibitem[DLAH24]{decatur20243d}
\textsc{Decatur D., Lang I., Aberman K., Hanocka R.}:
\newblock 3d paintbrush: Local stylization of 3d shapes with cascaded score distillation.
\newblock In \emph{Proceedings of the IEEE/CVF Conference on Computer Vision and Pattern Recognition} (2024), pp.~4473--4483.

\bibitem[DLLZ22]{du2022survey}
\textsc{Du Y., Liu Z., Li J., Zhao W.~X.}:
\newblock A survey of vision-language pre-trained models.
\newblock \emph{arXiv preprint arXiv:2202.10936} (2022).

\bibitem[DXY{\etalchar{*}}21]{dong2021location}
\textsc{Dong Z., Xu K., Yang Y., Bao H., Xu W., Lau R.~W.}:
\newblock Location-aware single image reflection removal.
\newblock In \emph{Proceedings of the IEEE/CVF International Conference on Computer Vision} (2021), pp.~5017--5026.

\bibitem[ERO21]{esser2021taming}
\textsc{Esser P., Rombach R., Ommer B.}:
\newblock Taming transformers for high-resolution image synthesis.
\newblock In \emph{Proceedings of the IEEE/CVF conference on computer vision and pattern recognition} (2021), pp.~12873--12883.

\bibitem[FLNP{\etalchar{*}}24]{fischer2024nerf}
\textsc{Fischer M., Li Z., Nguyen-Phuoc T., Bozic A., Dong Z., Marshall C., Ritschel T.}:
\newblock Nerf analogies: Example-based visual attribute transfer for nerfs.
\newblock In \emph{Proceedings of the IEEE/CVF Conference on Computer Vision and Pattern Recognition} (2024), pp.~4640--4650.

\bibitem[FWX{\etalchar{*}}22]{fang2022eva}
\textsc{Fang Y., Wang W., Xie B., Sun Q., Wu L., Wang X., Huang T., Wang X., Cao Y.}:
\newblock Eva: Exploring the limits of masked visual representation learning at scale.
\newblock \emph{arXiv preprint arXiv:2211.07636} (2022).

\bibitem[GAG{\etalchar{*}}23]{gao2023textdeformer}
\textsc{Gao W., Aigerman N., Groueix T., Kim V., Hanocka R.}:
\newblock Textdeformer: Geometry manipulation using text guidance.
\newblock In \emph{ACM SIGGRAPH 2023 Conference Proceedings} (2023), pp.~1--11.

\bibitem[GLKC21]{gu2021open}
\textsc{Gu X., Lin T.-Y., Kuo W., Cui Y.}:
\newblock Open-vocabulary object detection via vision and language knowledge distillation.
\newblock \emph{arXiv preprint arXiv:2104.13921} (2021).

\bibitem[GLS{\etalchar{*}}23]{threestudio2023}
\textsc{Guo Y.-C., Liu Y.-T., Shao R., Laforte C., Voleti V., Luo G., Chen C.-H., Zou Z.-X., Wang C., Cao Y.-P., Zhang S.-H.}:
\newblock threestudio: A unified framework for 3d content generation, 2023.
\newblock URL: \url{https://github.com/threestudio-project/threestudio}.

\bibitem[GLZ{\etalchar{*}}21]{guo2021volumetric}
\textsc{Guo J., Li M., Zong Z., Liu Y., He J., Guo Y., Yan L.-Q.}:
\newblock Volumetric appearance stylization with stylizing kernel prediction network.
\newblock \emph{ACM Trans. Graph. 40}, 4 (2021), 162--1.

\bibitem[GPM{\etalchar{*}}22]{gal2022stylegan}
\textsc{Gal R., Patashnik O., Maron H., Bermano A.~H., Chechik G., Cohen-Or D.}:
\newblock Stylegan-nada: Clip-guided domain adaptation of image generators.
\newblock \emph{ACM Transactions on Graphics (TOG) 41}, 4 (2022), 1--13.

\bibitem[HHY{\etalchar{*}}22]{huang2022stylizednerf}
\textsc{Huang Y.-H., He Y., Yuan Y.-J., Lai Y.-K., Gao L.}:
\newblock Stylizednerf: consistent 3d scene stylization as stylized nerf via 2d-3d mutual learning.
\newblock In \emph{Proceedings of the IEEE/CVF Conference on Computer Vision and Pattern Recognition} (2022), pp.~18342--18352.

\bibitem[HJN22]{hollein2022stylemesh}
\textsc{H{\"o}llein L., Johnson J., Nie{\ss}ner M.}:
\newblock Stylemesh: Style transfer for indoor 3d scene reconstructions.
\newblock In \emph{Proceedings of the IEEE/CVF Conference on Computer Vision and Pattern Recognition} (2022), pp.~6198--6208.

\bibitem[HKK23]{hwang2023text2scene}
\textsc{Hwang I., Kim H., Kim Y.~M.}:
\newblock Text2scene: Text-driven indoor scene stylization with part-aware details.
\newblock In \emph{Proceedings of the IEEE/CVF Conference on Computer Vision and Pattern Recognition} (2023), pp.~1890--1899.

\bibitem[HLG{\etalchar{*}}22]{hu2022subdivision}
\textsc{Hu S.-M., Liu Z.-N., Guo M.-H., Cai J.-X., Huang J., Mu T.-J., Martin R.~R.}:
\newblock Subdivision-based mesh convolution networks.
\newblock \emph{ACM Transactions on Graphics (TOG) 41}, 3 (2022), 1--16.

\bibitem[HTE{\etalchar{*}}23]{haque2023instruct}
\textsc{Haque A., Tancik M., Efros A.~A., Holynski A., Kanazawa A.}:
\newblock Instruct-nerf2nerf: Editing 3d scenes with instructions.
\newblock In \emph{Proceedings of the IEEE/CVF International Conference on Computer Vision} (2023), pp.~19740--19750.

\bibitem[HTS{\etalchar{*}}21]{huang2021learning}
\textsc{Huang H.-P., Tseng H.-Y., Saini S., Singh M., Yang M.-H.}:
\newblock Learning to stylize novel views.
\newblock In \emph{Proceedings of the IEEE/CVF International Conference on Computer Vision} (2021), pp.~13869--13878.

\bibitem[HZP{\etalchar{*}}22]{hong2022avatarclip}
\textsc{Hong F., Zhang M., Pan L., Cai Z., Yang L., Liu Z.}:
\newblock Avatarclip: Zero-shot text-driven generation and animation of 3d avatars.
\newblock \emph{arXiv preprint arXiv:2205.08535} (2022).

\bibitem[JNS{\etalchar{*}}24]{jung2024geometry}
\textsc{Jung H., Nam S., Sarafianos N., Yoo S., Sorkine-Hornung A., Ranjan R.}:
\newblock Geometry transfer for stylizing radiance fields.
\newblock In \emph{Proceedings of the IEEE/CVF Conference on Computer Vision and Pattern Recognition} (2024), pp.~8565--8575.

\bibitem[JYF{\etalchar{*}}19]{jing2019neural}
\textsc{Jing Y., Yang Y., Feng Z., Ye J., Yu Y., Song M.}:
\newblock Neural style transfer: A review.
\newblock \emph{IEEE transactions on visualization and computer graphics 26}, 11 (2019), 3365--3385.

\bibitem[KAGS19]{kim2019transport}
\textsc{Kim B., Azevedo V.~C., Gross M., Solenthaler B.}:
\newblock Transport-based neural style transfer for smoke simulations.
\newblock \emph{arXiv preprint arXiv:1905.07442} (2019).

\bibitem[KAGS20]{kim2020lagrangian}
\textsc{Kim B., Azevedo V.~C., Gross M., Solenthaler B.}:
\newblock Lagrangian neural style transfer for fluids.
\newblock \emph{ACM Transactions on Graphics (TOG) 39}, 4 (2020), 52--1.

\bibitem[KRA{\etalchar{*}}20]{kuznetsova2020open}
\textsc{Kuznetsova A., Rom H., Alldrin N., Uijlings J., Krasin I., Pont-Tuset J., Kamali S., Popov S., Malloci M., Kolesnikov A., et~al.}:
\newblock The open images dataset v4: Unified image classification, object detection, and visual relationship detection at scale.
\newblock \emph{International Journal of Computer Vision 128}, 7 (2020), 1956--1981.

\bibitem[KSL{\etalchar{*}}21]{kamath2021mdetr}
\textsc{Kamath A., Singh M., LeCun Y., Synnaeve G., Misra I., Carion N.}:
\newblock Mdetr-modulated detection for end-to-end multi-modal understanding.
\newblock In \emph{Proceedings of the IEEE/CVF International Conference on Computer Vision} (2021), pp.~1780--1790.

\bibitem[KUH18]{kato2018neural}
\textsc{Kato H., Ushiku Y., Harada T.}:
\newblock Neural 3d mesh renderer.
\newblock In \emph{Proceedings of the IEEE conference on computer vision and pattern recognition} (2018), pp.~3907--3916.

\bibitem[KY22]{kwon2022clipstyler}
\textsc{Kwon G., Ye J.~C.}:
\newblock Clipstyler: Image style transfer with a single text condition.
\newblock In \emph{Proceedings of the IEEE/CVF Conference on Computer Vision and Pattern Recognition} (2022), pp.~18062--18071.

\bibitem[LDG{\etalchar{*}}17]{lin2017feature}
\textsc{Lin T.-Y., Doll{\'a}r P., Girshick R., He K., Hariharan B., Belongie S.}:
\newblock Feature pyramid networks for object detection.
\newblock In \emph{Proceedings of the IEEE conference on computer vision and pattern recognition} (2017), pp.~2117--2125.

\bibitem[LGG{\etalchar{*}}17]{lin2017focal}
\textsc{Lin T.-Y., Goyal P., Girshick R., He K., Doll{\'a}r P.}:
\newblock Focal loss for dense object detection.
\newblock In \emph{Proceedings of the IEEE international conference on computer vision} (2017), pp.~2980--2988.

\bibitem[LGT{\etalchar{*}}23]{lin2023magic3d}
\textsc{Lin C.-H., Gao J., Tang L., Takikawa T., Zeng X., Huang X., Kreis K., Fidler S., Liu M.-Y., Lin T.-Y.}:
\newblock Magic3d: High-resolution text-to-3d content creation.
\newblock In \emph{Proceedings of the IEEE/CVF Conference on Computer Vision and Pattern Recognition} (2023), pp.~300--309.

\bibitem[LLW{\etalchar{*}}23]{li2023gligen}
\textsc{Li Y., Liu H., Wu Q., Mu F., Yang J., Gao J., Li C., Lee Y.~J.}:
\newblock Gligen: Open-set grounded text-to-image generation.
\newblock In \emph{Proceedings of the IEEE/CVF Conference on Computer Vision and Pattern Recognition} (2023), pp.~22511--22521.

\bibitem[LMB{\etalchar{*}}14]{lin2014microsoft}
\textsc{Lin T.-Y., Maire M., Belongie S., Hays J., Perona P., Ramanan D., Doll{\'a}r P., Zitnick C.~L.}:
\newblock Microsoft coco: Common objects in context.
\newblock In \emph{Computer Vision--ECCV 2014: 13th European Conference, Zurich, Switzerland, September 6-12, 2014, Proceedings, Part V 13} (2014), Springer, pp.~740--755.

\bibitem[LSG{\etalchar{*}}21]{li2021align}
\textsc{Li J., Selvaraju R., Gotmare A., Joty S., Xiong C., Hoi S. C.~H.}:
\newblock Align before fuse: Vision and language representation learning with momentum distillation.
\newblock \emph{Advances in neural information processing systems 34} (2021), 9694--9705.

\bibitem[LTJ18]{liu2018paparazzi}
\textsc{Liu H.-T.~D., Tao M., Jacobson A.}:
\newblock Paparazzi: surface editing by way of multi-view image processing.
\newblock \emph{ACM Trans. Graph. 37}, 6 (2018), 221--1.

\bibitem[LYL{\etalchar{*}}20]{li2020oscar}
\textsc{Li X., Yin X., Li C., Zhang P., Hu X., Zhang L., Wang L., Hu H., Dong L., Wei F., et~al.}:
\newblock Oscar: Object-semantics aligned pre-training for vision-language tasks.
\newblock In \emph{Computer Vision--ECCV 2020: 16th European Conference, Glasgow, UK, August 23--28, 2020, Proceedings, Part XXX 16} (2020), Springer, pp.~121--137.

\bibitem[LYY{\etalchar{*}}19]{li2019visualbert}
\textsc{Li L.~H., Yatskar M., Yin D., Hsieh C.-J., Chang K.-W.}:
\newblock Visualbert: A simple and performant baseline for vision and language.
\newblock \emph{arXiv preprint arXiv:1908.03557} (2019).

\bibitem[LZC{\etalchar{*}}23]{liu2023stylerf}
\textsc{Liu K., Zhan F., Chen Y., Zhang J., Yu Y., El~Saddik A., Lu S., Xing E.~P.}:
\newblock Stylerf: Zero-shot 3d style transfer of neural radiance fields.
\newblock In \emph{Proceedings of the IEEE/CVF Conference on Computer Vision and Pattern Recognition} (2023), pp.~8338--8348.

\bibitem[LZZ{\etalchar{*}}22]{li2022grounded}
\textsc{Li L.~H., Zhang P., Zhang H., Yang J., Li C., Zhong Y., Wang L., Yuan L., Zhang L., Hwang J.-N., et~al.}:
\newblock Grounded language-image pre-training.
\newblock In \emph{Proceedings of the IEEE/CVF Conference on Computer Vision and Pattern Recognition} (2022), pp.~10965--10975.

\bibitem[MBOL{\etalchar{*}}22]{michel2022text2mesh}
\textsc{Michel O., Bar-On R., Liu R., Benaim S., Hanocka R.}:
\newblock Text2mesh: Text-driven neural stylization for meshes.
\newblock In \emph{Proceedings of the IEEE/CVF Conference on Computer Vision and Pattern Recognition} (2022), pp.~13492--13502.

\bibitem[MGA{\etalchar{*}}17]{maron2017convolutional}
\textsc{Maron H., Galun M., Aigerman N., Trope M., Dym N., Yumer E., Kim V.~G., Lipman Y.}:
\newblock Convolutional neural networks on surfaces via seamless toric covers.
\newblock \emph{ACM Trans. Graph. 36}, 4 (2017), 71--1.

\bibitem[MKXBP22]{mohammad2022clip}
\textsc{Mohammad~Khalid N., Xie T., Belilovsky E., Popa T.}:
\newblock Clip-mesh: Generating textured meshes from text using pretrained image-text models.
\newblock In \emph{SIGGRAPH Asia 2022 Conference Papers} (2022), pp.~1--8.

\bibitem[MRP{\etalchar{*}}23]{metzer2023latent}
\textsc{Metzer G., Richardson E., Patashnik O., Giryes R., Cohen-Or D.}:
\newblock Latent-nerf for shape-guided generation of 3d shapes and textures.
\newblock In \emph{Proceedings of the IEEE/CVF Conference on Computer Vision and Pattern Recognition} (2023), pp.~12663--12673.

\bibitem[MST{\etalchar{*}}21]{mildenhall2021nerf}
\textsc{Mildenhall B., Srinivasan P.~P., Tancik M., Barron J.~T., Ramamoorthi R., Ng R.}:
\newblock Nerf: Representing scenes as neural radiance fields for view synthesis.
\newblock \emph{Communications of the ACM 65}, 1 (2021), 99--106.

\bibitem[NDR{\etalchar{*}}21]{nichol2021glide}
\textsc{Nichol A., Dhariwal P., Ramesh A., Shyam P., Mishkin P., McGrew B., Sutskever I., Chen M.}:
\newblock Glide: Towards photorealistic image generation and editing with text-guided diffusion models.
\newblock \emph{arXiv preprint arXiv:2112.10741} (2021).

\bibitem[NPLX22]{nguyen2022snerf}
\textsc{Nguyen-Phuoc T., Liu F., Xiao L.}:
\newblock Snerf: stylized neural implicit representations for 3d scenes.
\newblock \emph{arXiv preprint arXiv:2207.02363} (2022).

\bibitem[PJBM23]{poole2023dreamfusion}
\textsc{Poole B., Jain A., Barron J.~T., Mildenhall B.}:
\newblock Dreamfusion: Text-to-3d using 2d diffusion.
\newblock In \emph{The Eleventh International Conference on Learning Representations} (2023).
\newblock URL: \url{https://openreview.net/forum?id=FjNys5c7VyY}.

\bibitem[PL22]{pinkney2022clip2latent}
\textsc{Pinkney J.~N., Li C.}:
\newblock clip2latent: Text driven sampling of a pre-trained stylegan using denoising diffusion and clip.
\newblock \emph{arXiv preprint arXiv:2210.02347} (2022).

\bibitem[RBL{\etalchar{*}}22]{rombach2022high}
\textsc{Rombach R., Blattmann A., Lorenz D., Esser P., Ommer B.}:
\newblock High-resolution image synthesis with latent diffusion models.
\newblock In \emph{Proceedings of the IEEE/CVF conference on computer vision and pattern recognition} (2022), pp.~10684--10695.

\bibitem[RDN{\etalchar{*}}22]{ramesh2022hierarchical}
\textsc{Ramesh A., Dhariwal P., Nichol A., Chu C., Chen M.}:
\newblock Hierarchical text-conditional image generation with clip latents.
\newblock \emph{arXiv preprint arXiv:2204.06125} (2022).

\bibitem[Reh21]{Toys4K}
\textsc{Rehg J.~M.}:
\newblock Toys4k 3d object dataset, 2021.
\newblock URL: \url{https://github.com/rehg-lab/lowshot-shapebias/tree/main/toys4k}.

\bibitem[RKH{\etalchar{*}}21]{radford2021learning}
\textsc{Radford A., Kim J.~W., Hallacy C., Ramesh A., Goh G., Agarwal S., Sastry G., Askell A., Mishkin P., Clark J., et~al.}:
\newblock Learning transferable visual models from natural language supervision.
\newblock In \emph{International conference on machine learning} (2021), PMLR, pp.~8748--8763.

\bibitem[RMA{\etalchar{*}}23]{richardson2023texture}
\textsc{Richardson E., Metzer G., Alaluf Y., Giryes R., Cohen-Or D.}:
\newblock Texture: Text-guided texturing of 3d shapes.
\newblock \emph{arXiv preprint arXiv:2302.01721} (2023).

\bibitem[SHG{\etalchar{*}}22]{singh2022flava}
\textsc{Singh A., Hu R., Goswami V., Couairon G., Galuba W., Rohrbach M., Kiela D.}:
\newblock Flava: A foundational language and vision alignment model.
\newblock In \emph{Proceedings of the IEEE/CVF Conference on Computer Vision and Pattern Recognition} (2022), pp.~15638--15650.

\bibitem[SLZ{\etalchar{*}}19]{shao2019objects365}
\textsc{Shao S., Li Z., Zhang T., Peng C., Yu G., Zhang X., Li J., Sun J.}:
\newblock Objects365: A large-scale, high-quality dataset for object detection.
\newblock In \emph{Proceedings of the IEEE/CVF international conference on computer vision} (2019), pp.~8430--8439.

\bibitem[SZ14]{simonyan2014very}
\textsc{Simonyan K., Zisserman A.}:
\newblock Very deep convolutional networks for large-scale image recognition.
\newblock \emph{arXiv preprint arXiv:1409.1556} (2014).

\bibitem[TMT{\etalchar{*}}23]{tsalicoglou2023textmesh}
\textsc{Tsalicoglou C., Manhardt F., Tonioni A., Niemeyer M., Tombari F.}:
\newblock Textmesh: Generation of realistic 3d meshes from text prompts.
\newblock \emph{arXiv preprint arXiv:2304.12439} (2023).

\bibitem[TSCH20]{tian2020fcos}
\textsc{Tian Z., Shen C., Chen H., He T.}:
\newblock Fcos: A simple and strong anchor-free object detector.
\newblock \emph{IEEE Transactions on Pattern Analysis and Machine Intelligence 44}, 4 (2020), 1922--1933.

\bibitem[{Tur}22]{turbosquid2022}
\textsc{{TurboSquid}}:
\newblock Turbosquid 3d model repository, 2022.
\newblock URL: \url{https://www.turbosquid.com/}.

\bibitem[WAVK{\etalchar{*}}12]{wang2012active}
\textsc{Wang Y., Asafi S., Van~Kaick O., Zhang H., Cohen-Or D., Chen B.}:
\newblock Active co-analysis of a set of shapes.
\newblock \emph{ACM Transactions on Graphics (TOG) 31}, 6 (2012), 1--10.

\bibitem[WBL22]{wang2022yolov7}
\textsc{Wang C.-Y., Bochkovskiy A., Liao H.-Y.~M.}:
\newblock Yolov7: Trainable bag-of-freebies sets new state-of-the-art for real-time object detectors.
\newblock \emph{arXiv preprint arXiv:2207.02696} (2022).

\bibitem[WDC{\etalchar{*}}22]{wang2022internimage}
\textsc{Wang W., Dai J., Chen Z., Huang Z., Li Z., Zhu X., Hu X., Lu T., Lu L., Li H., et~al.}:
\newblock Internimage: Exploring large-scale vision foundation models with deformable convolutions.
\newblock \emph{arXiv preprint arXiv:2211.05778} (2022).

\bibitem[WLW{\etalchar{*}}23]{wang2023prolificdreamer}
\textsc{Wang Z., Lu C., Wang Y., Bao F., Li C., Su H., Zhu J.}:
\newblock Prolificdreamer: High-fidelity and diverse text-to-3d generation with variational score distillation.
\newblock \emph{arXiv preprint arXiv:2305.16213} (2023).

\bibitem[WSK{\etalchar{*}}15]{wu20153d}
\textsc{Wu Z., Song S., Khosla A., Yu F., Zhang L., Tang X., Xiao J.}:
\newblock 3d shapenets: A deep representation for volumetric shapes.
\newblock In \emph{Proceedings of the IEEE conference on computer vision and pattern recognition} (2015), pp.~1912--1920.

\bibitem[XSS23]{xu2023stylerdalle}
\textsc{Xu Z., Sangineto E., Sebe N.}:
\newblock Stylerdalle: Language-guided style transfer using a vector-quantized tokenizer of a large-scale generative model.
\newblock \emph{arXiv preprint arXiv:2303.09268} (2023).

\bibitem[XWL{\etalchar{*}}22]{xu2022pp}
\textsc{Xu S., Wang X., Lv W., Chang Q., Cui C., Deng K., Wang G., Dang Q., Wei S., Du Y., et~al.}:
\newblock Pp-yoloe: An evolved version of yolo.
\newblock \emph{arXiv preprint arXiv:2203.16250} (2022).

\bibitem[YCP{\etalchar{*}}23]{yang20233dstyle}
\textsc{Yang H., Chen Y., Pan Y., Yao T., Chen Z., Mei T.}:
\newblock 3dstyle-diffusion: Pursuing fine-grained text-driven 3d stylization with 2d diffusion models.
\newblock In \emph{Proceedings of the 31st ACM International Conference on Multimedia} (2023), pp.~6860--6868.

\bibitem[YLK{\etalchar{*}}21]{yu2021vector}
\textsc{Yu J., Li X., Koh J.~Y., Zhang H., Pang R., Qin J., Ku A., Xu Y., Baldridge J., Wu Y.}:
\newblock Vector-quantized image modeling with improved vqgan.
\newblock \emph{arXiv preprint arXiv:2110.04627} (2021).

\bibitem[ZCY{\etalchar{*}}20]{zhang2020bridging}
\textsc{Zhang S., Chi C., Yao Y., Lei Z., Li S.~Z.}:
\newblock Bridging the gap between anchor-based and anchor-free detection via adaptive training sample selection.
\newblock In \emph{Proceedings of the IEEE/CVF conference on computer vision and pattern recognition} (2020), pp.~9759--9768.

\bibitem[ZWL{\etalchar{*}}23]{zhuang2023dreameditor}
\textsc{Zhuang J., Wang C., Lin L., Liu L., Li G.}:
\newblock Dreameditor: Text-driven 3d scene editing with neural fields.
\newblock In \emph{SIGGRAPH Asia 2023 Conference Papers} (2023), pp.~1--10.

\bibitem[ZZH{\etalchar{*}}22]{zhang2022glipv2}
\textsc{Zhang H., Zhang P., Hu X., Chen Y.-C., Li L., Dai X., Wang L., Yuan L., Hwang J.-N., Gao J.}:
\newblock Glipv2: Unifying localization and vision-language understanding.
\newblock \emph{Advances in Neural Information Processing Systems 35} (2022), 36067--36080.

\bibitem[ZZK23]{zhu2023hifa}
\textsc{Zhu J., Zhuang P., Koyejo S.}:
\newblock Hifa: High-fidelity text-to-3d generation with advanced diffusion guidance.
\newblock In \emph{The Twelfth International Conference on Learning Representations} (2023).

\end{thebibliography}


\end{document}